\documentclass{article}
\pdfoutput=1
\usepackage{spconf,amsmath,graphicx} 

\usepackage{enumitem}
\setlist{nosep, leftmargin=14pt}

\usepackage{mwe} 
\usepackage{booktabs} 
\usepackage{multirow}
\usepackage{tabularx}
\usepackage{adjustbox}
\usepackage{array}
\usepackage{svg}
\usepackage{gensymb}
\usepackage{anyfontsize}
\usepackage{amssymb}
\usepackage{pifont}
\usepackage{algorithm,algorithmicx}
\usepackage[noend]{algpseudocode}

\usepackage{makecell}

\title{Boosting Medical Image Classification with \\ Segmentation Foundation Model}
\vspace{-6mm}
\name{\normalsize{Pengfei Gu$^\dagger$ 
      Zihan Zhao$^\diamond$ 
      Hongxiao Wang$^\dagger$ 
      Yaopeng Peng$^\dagger$
      Yizhe Zhang$^\star$ \thanks{Corresponding author: Yizhe Zhang}
      Nishchal Sapkota$^\dagger$
      Chaoli Wang$^\dagger$ 
      Danny Z. Chen$^\dagger$}}

\address{\normalsize{
$^\star$Nanjing University of Science and Technology, Nanjing, Jiangsu 210094, China} \\
\normalsize{$^\dagger$University of Notre Dame, Notre Dame, IN 46556, USA}
\\
\normalsize{$^\diamond$Tianjin University, Tianjin, Tianjin 300072, China}}

\begin{document}
%
\maketitle
\begin{abstract}
The Segment Anything Model (SAM) exhibits impressive capabilities in zero-shot segmentation for natural images.
Recently, SAM has gained a great deal of attention for its applications in medical image segmentation.
However, to our best knowledge, no studies have shown how to harness the power of SAM for medical image classification.
To fill this gap and make SAM a true ``foundation model'' for medical image analysis, it is highly desirable to customize SAM specifically for medical image classification.
In this paper, we introduce SAMAug-C, an innovative augmentation method based on SAM for augmenting classification datasets by generating variants of the original images.
The augmented datasets can be used to train a deep learning classification model, thereby boosting the classification performance.
Furthermore, we propose a novel framework that simultaneously processes raw and SAMAug-C augmented image input, capitalizing on the complementary information that is offered by both.
Experiments on three public datasets validate the effectiveness of our new approach.
\end{abstract}
%
%
\section{Introduction} \label{intro}
\vspace{-3mm}
Trained on over 1 billion tasks using 11 million images, the Segment Anything Model (SAM)~\cite{kirillov2023segment}, a Segmentation Foundation Model, has showcased impressive zero-shot image segmentation capabilities for natural images across various prompts, such as points, boxes, and masks.
Recently, a number of studies have explored leveraging SAM for medical image segmentation, either by directly applying  SAM~\cite{ deng2023segment,mattjie2023exploring,huang2023segment,zhang2023input} or by fine-tuning SAM for medical images
\cite{ma2023segment,li2023polyp,zhang2023customized}. 
Despite these efforts, some studies have exhibited unsatisfactory performance of using SAM in medical image segmentation~\cite{ma2023segment,deng2023segment} due to the following challenges: (1) the large differences in appearance between medical and natural images, and (2) the often blurred boundaries of target objects in medical images.
\begin{figure}[t!]
    \centering
    \includegraphics[width=0.42\textwidth]{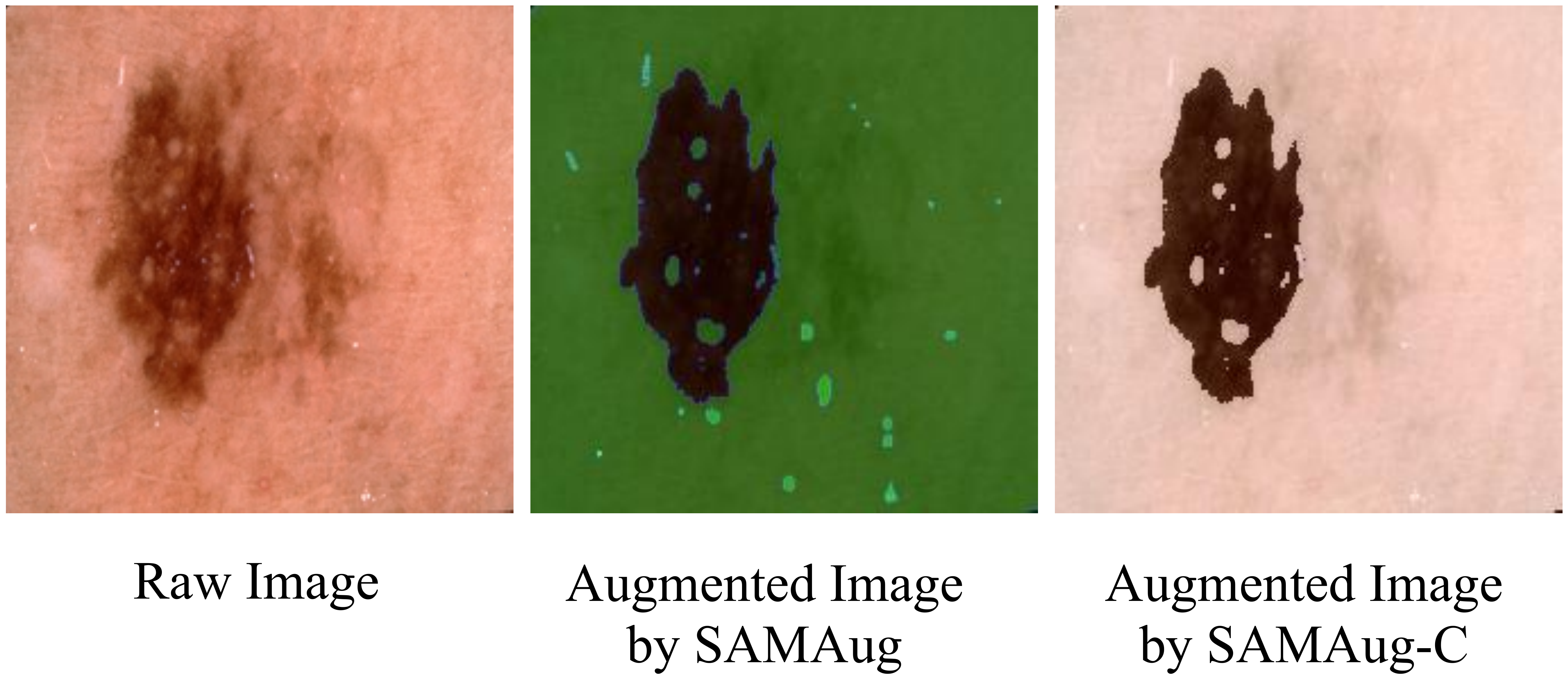}
    \vspace*{-0.15in}
    \caption{Visual example from the ISIC 2017 dataset~\cite{codella2018skin}. The entire image  augmented by SAMAug~\cite{zhang2023input} is covered in green, which may increase the difficulty for a classifier to distinguish the skin lesion from the background.}
    \label{fig:visual}
\vspace*{-4mm}
\end{figure}
%
In contrast to the SAM-based methods mentioned above for medical image segmentation, SAMAug~\cite{zhang2023input} employed SAM to augment raw image input for commonly-used deep learning (DL) medical image segmentation models (e.g., U-Net~\cite{ronneberger2015u}), thereby enhancing their segmentation performances.
Specifically, SAMAug augments raw images with segmentation maps and boundary prior maps generated by SAM (i.e., adding the segmentation and boundary prior maps to the second and third channels of the raw images, respectively). 

Medical image classification is a pivotal task in diagnostic medicine, assisting clinicians in their decision-making processes~\cite{chen2020pathomic}.
To our best knowledge, SAM has yet to be employed for medical image classification.
While one might consider using SAMAug, inspired by its capabilities, to augment raw image input for classification, several challenges arise.
First, SAMAug was primarily designed to augment raw images for medical image segmentation, emphasizing the use of both SAM-generated segmentation maps and boundary prior maps.
Our experiments have shown that applying SAMAug directly to raw images intended for classification could lead to a performance drop.
As evidence, in Table~\ref{tab:intro}, we observe performance drops on the ISIC 2017 skin lesion classification dataset~\cite{codella2018skin} when training two DL-based classification models (ResNet152~\cite{he2016deep} and SENet154~\cite{hu2018squeeze}) with raw images augmented by SAMAug.
Second, the SAM-generated segmentation maps and boundary prior maps could inadvertently obscure crucial regions in the raw images. Visual example of such issues with augmented image produced by SAMAug, using sample from the ISIC 2017 skin lesion classification dataset, is illustrated in Fig.~\ref{fig:visual}.

With these challenges in mind, it becomes imperative to customize SAM specifically for medical image classification.
This paper addresses two pivotal questions for medical image classification: (I) How can we design an effective SAM-based augmentation method that emphasizes the crucial regions while suppressing irrelevant ones in input images for medical image classification?
(II) How can we effectively utilize raw and SAM-augmented images to enhance classification performance?

For the first question, we introduce SAMAug-C, an innovative augmentation method built upon SAM, aiming to augment the input datasets by creating variations of the original images.
Initially, we leverage SAM's zero-shot image segmentation capability to procure segmentation masks for the raw images. We then generate the corresponding segmentation prior maps by assigning a value of `1' to the masked regions.
Subsequently, these segmentation prior maps are added to each channel of the raw images to augment them.

For the second question, we propose a novel framework that processes both the raw and SAMAug-C augmented images simultaneously, harnessing the complementary information that each of them provides.
The framework consists of two branches, both equipped with identical backbone models.
These models are concurrently trained with the raw and SAMAug-C augmented images.
Subsequently, an ensemble module is employed to amalgamate the predictions from the two branches, yielding the final predicted label.

In summary, our main contributions are as follows: (1) We adapt SAM specifically for medical image classification, pioneering its use in this domain. 
(2) We present SAMAug-C, which is designed to augment raw images for medical image classification, and put forward a new framework that effectively trains on both raw and SAMAug-C augmented images simultaneously.
(3) We conduct comprehensive experiments on three public medical image classification datasets to demonstrate the effectiveness of our new approach.

\begin{table}[t!]
    \centering
    \caption{Results on the ISIC 2017 dataset. 
    }
    \label{tab:intro}
    \scriptsize
    \resizebox{\columnwidth}{!}
    {
    \begin{tabular}{l|l|l|l|l}
        \hline
        Method &Acc ($\uparrow$)  &AUC ($\uparrow$) &Sen ($\uparrow$)  &Spe ($\uparrow$)\\\hline
        ResNet152~\cite{he2016deep}&\textbf{84.53}	&\textbf{81.28}	&\textbf{49.23}	&\textbf{93.08}\\
         ResNet152 + SAMAug~\cite{zhang2023input}& 82.13	&76.01	&37.78	&92.88\\\hline
         SENet154~\cite{hu2018squeeze}& \textbf{84.45}	&\textbf{79.41}	&\textbf{42.74}	&\textbf{94.53}\\
        SENet154 + SAMAug~\cite{zhang2023input}& 82.53	&77.00	&37.09	&93.54\\\hline
    \end{tabular}
}
\vspace{-6mm}
\end{table}

\vspace{-4mm}
\section{Methodology} \label{method}
\vspace{-3mm}

\subsection{Background: The SAM Architecture} \label{background}
\vspace{-2mm}

SAM consists of three main components: an image encoder, a prompt encoder, and a mask decoder.
%
The image encoder accepts an input image of any size and produces an embedding feature.
The prompt encoder can handle both sparse prompts (e.g., boxes) and dense prompts (e.g., masks).
The mask decoder is a Transformer decoder block modified to incorporate a dynamic mask prediction head.
SAM employs a two-way attention module, one for prompt-to-image embedding and the other for image-to-prompt embedding in each block, facilitating learning of the interactions between the prompt and image embeddings.
After processing through two blocks, SAM upsamples the image embedding. Then, a MLP maps the output token to a dynamic linear classifier, which then predicts the target mask for the provided image.
In this paper, we leverage SAM to predict the mask of the input image.

\vspace{-4mm}
\subsection{SAMAug-C: Augmenting Input Raw Images for Medical Image Classification} \label{SAMAug-C}
\vspace{-2mm}
For a given raw image $I$, SAMAug~\cite{zhang2023input} produces two corresponding prior maps for $I$. The first one is a segmentation prior map derived from the mask's stability score generated by SAM. The second one is a boundary prior map, representing the exterior boundary of the segmentation mask. 
The raw image $I$ is augmented by adding to $I$ the segmentation prior map to its second channel and the boundary prior map to its third channel.
While this augmentation strategy was shown to be effective for several medical image segmentation tasks~\cite{zhang2023input}, it falls short in performance for medical image classification (e.g., as shown in Table \ref{tab:intro}).

To address this limitation and develop a more effective augmentation method for highlighting the important regions and suppressing irrelevant ones in the input images for medical image classification, we introduce SAMAug-C, augmenting raw image input for medical image classification.

As illustrated in Algorithm~\ref{algorithm1}, for a given input raw image, SAM's mask generator first predicts segmentation masks and stores them in a list.
For every segmentation mask in this list, we generate a corresponding segmentation prior map, assigning a value of 1 to the masked region.
Then, we combine all the segmentation prior maps to produce a final segmentation prior map for the raw image.
We then set the values of all the masked regions in this final segmentation prior map to 1, and augment the raw image by adding the segmentation prior map to each channel of the raw image.
It is important to note that if SAM does not generate any segmentation masks, then the raw image remains unaugmented.
The added segmentation prior map effectively emphasizes crucial regions and diminishes irrelevant ones in the input image. Refer to Fig.~\ref{fig:visual} for visual example of image augmented using SAMAug-C.

\begin{algorithm}[t]
\caption{{SAMAug-C}($tI, mask\_generator$)}\label{algorithm1}
\begin{algorithmic}[1]
    \State {Input: The raw image input $tI$ and mask generator $mask\_generator$ from SAM}
    \State {Output: Augmented image $newTI$}
    \State $masks \gets mask\_generator.\Call{generate}{tI}$
    \State $tI \gets \Call{img\_as\_float}{tI}$
    \State $SegPrior \gets \Call{ZerosMatrix}{\text{size of } tI[0], tI[1]}$
    \State $newTI \gets \Call{ZerosMatrix}{\text{size of } tI[0], tI[1], tI[2]}$

    \For{$maskindex$ \textbf{from} $0$ \textbf{to} $\text{length of } masks - 1$}
        \State $thismask \gets masks[maskindex]['segmentation']$
        \State $thismask\_ \gets \Call{ZerosMatrix}{\text{size of } thismask}$
        \State $thismask\_[\text{where } thismask = \text{True}] \gets 1$
        \State $SegPrior[\text{where } thismask\_ = 1] \gets SegPrior[\text{where } thismask\_ = 1] + 1.0$
    \EndFor

    \If{$SegPrior.\Call{min}{} = SegPrior.\Call{max}{}$}
        \For{$i$ \textbf{from} $0$ \textbf{to} $2$}
            \State $newTI[:,:,i] \gets tI[:,:,i]$
        \EndFor
    \Else
        \State $SegPrior \gets \Call{Where}{SegPrior \neq 0, 1, 0}$
        \For{$i$ \textbf{from} $0$ \textbf{to} $2$}
            \State $newTI[:,:,i] \gets tI[:,:,i] + SegPrior$
        \EndFor
    \EndIf
    \Return $newTI$
\end{algorithmic}
\end{algorithm}
\vspace{-4mm}
\subsection{Model Training with Raw and SAMAug-C Augmented Images} \label{framework}
\vspace{-2mm}
Using SAMAug-C to augment input raw images, we derive a new set of images, called SAMAug-C augmented images.
This raises a pertinent question: How can we effectively leverage the SAMAug-C augmented images to enhance medical image classification?
A straightforward approach may involve: (I) During the training phase, employ SAMAug-C augmented images to train a DL classification model; 
(II) in the test phase, labels are predicted by feeding the model with SAMAug-C augmented test images.
We refer to this approach as ``DL classification model + SAMAug-C'' (e.g., ResNet152 + SAMAug-C).
While possibly effective, this straightforward design might struggle in scenarios where SAM fails to produce accurate segmentation masks.

\begin{figure}[t!]
    \centering
    \includegraphics[width=0.45\textwidth]{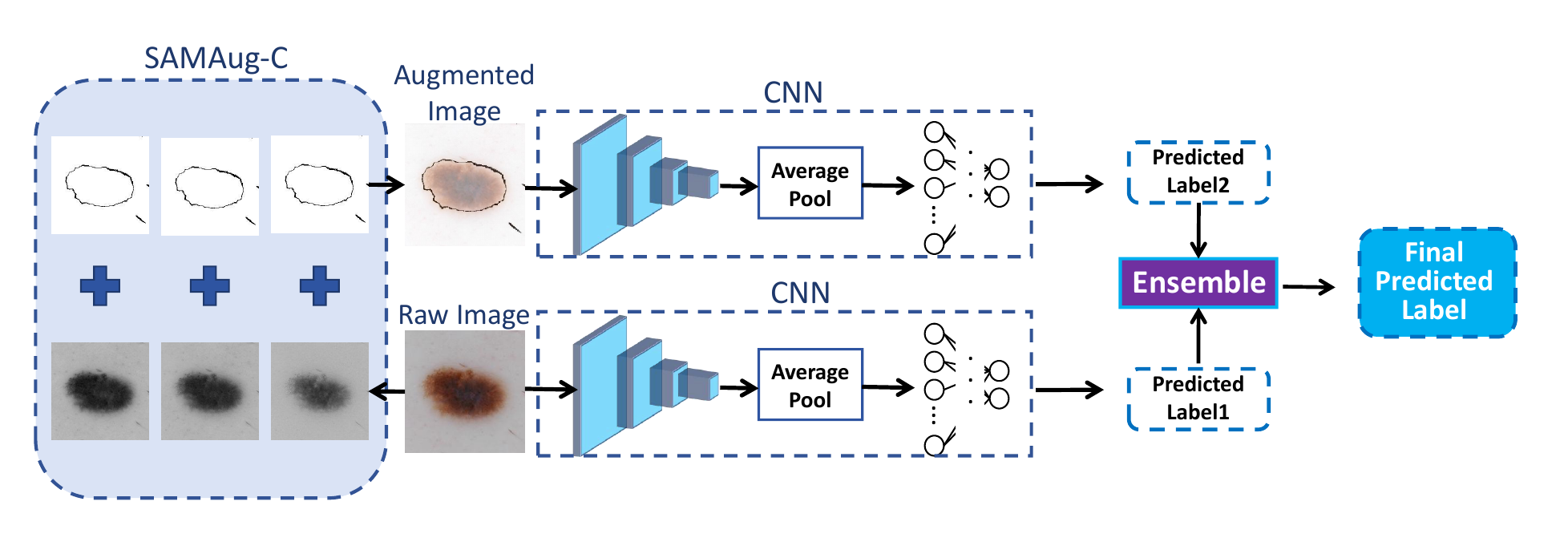}
    \vspace*{-0.25in}
    \caption{The overview of our framework.}
    \label{fig:framework}
\vspace*{-6mm}
\end{figure}

To address this limitation, we propose a novel framework that concurrently processes both the raw images and their SAMAug-C augmented images.
As shown in Fig.~\ref{fig:framework}, our new framework consists of two branches.
The ``Raw Image'' branch ingests raw image input to train a DL classification model (e.g., ResNet) and produces predicted labels. 
In contrast, the ``Augmented Image'' branch uses the augmented images generated by SAMAug-C to train another DL classification model, which shares the same architecture as that of the ``Raw Image'' branch, and subsequently outputs the associated predicted labels.
An ensemble module is employed to consolidate the outputs of the two branches, generating the final predicted label.
We refer to this design as ``DL classification model + SAMAug-C + Ensemble'' (e.g., ResNet152 + SAMAug-C + Ensemble).
\vspace{-4mm}
\subsection{Model Ensemble} \label{ensemble}
\vspace{-2mm}
For a given test image $x$ and its corresponding augmented test image generated by SAMAug-C, we employ two models to process them and ensemble the results.
From these two models, we obtain two $N$-dimensional vectors, $p_1$ and $p_2,$ where $N$ denotes the number of categories that the task is classified into.
We explore several possible ensemble schemes:
\textbf{(1) Voting}: The majority voting strategy is employed to derive the final prediction by consolidating the outcomes from both models.
\textbf{(2) Entropy}: The entropy offers a quantifiable measure of the uncertainty or randomness present in the predicted probabilities across various classes. We compute the entropy using the equation \(-p \times \log_2 p\), where \(p\) is the set of predicted probabilities. By multiplying each probability with its logarithm (base 2) and negating the result, we then accumulate these values across the classes to determine the entropy for each sample. Ultimately, the model output with the lower entropy is chosen from the two.
\textbf{(3) Direct average}: The outcomes from both models are averaged to arrive at the final prediction.
\textbf{(4) Weighted average}: During the averaging process, a weight, indicating the significance of each model's output, is employed:
$\text{Final prediction} = \frac{\sum_{i=1}^{2}\omega_i p_i}{2},$
where \(\omega_i\) represents the weight of the output of the \(i\)-th model, with \(\omega_i \geq 0\) and \(\omega_1 + \omega_2 = 1\). For clarity, we refer to the ``Raw Image'' branch as the \(1\)-st model and the ``Augmented Image'' branch as the \(2\)-nd model. 

Through empirical study, we choose the weighted average ensemble scheme for our framework in all the experiments.

\vspace{-4mm}
\section{Experiments and Results} \label{exp}
\label{sec:exp}
\vspace{-2mm}

\textbf{Datasets.}
\noindent
(1) The ISIC 2017 skin lesion classification dataset (ISIC 2017):
The dataset~\cite{codella2018skin} contains 2000 training, 150 validation, and 600 test images. Our experiments are focused on task-3A: melanoma detection.
(2) The vitiligo (public) dataset:
The dataset~\cite{zhang2021design} contains 672 training, 268 validation, and 401 test images. Our experiments are performed for vitiligo detection.
(3) The extended colorectal cancer (ExtCRC) grading dataset: 
The dataset~\cite{shaban2020context} contains 300 H\&E-stained colorectal cancer subtyping pathology images. The task is three categories (Grades 1, 2, and 3) classification.
We randomly split the data, allocating 80\% for training and 20\% for testing.
We resize all the images of each dataset to 224$\times$224. For the ISIC 2017 and vitiligo public datasets, our experiments conduct 5 runs using different seeds, and for the ExtCRC dataset, we perform the random data splitting for 5 times, presenting the average outcomes of our experiments.

\noindent
\textbf{Implementation Details.}
Our experiments are conducted using the PyTorch. The model is trained on an NVIDIA Tesla V100 Graphics Card (32GB GPU memory) using the AdamW optimizer with a weight decay = $0.005$. 
The learning rate is 0.0001, and the number of training epochs is 400 for the experiments. The batch size for each case is set as the maximum size allowed by the GPU.
Standard data augmentation (e.g., random flip, crop, etc.) is applied to avoid overfitting.

%

\begin{table}[t]
    \centering
    \caption{Results on the ISIC 2017 dataset. The best results are marked in {\bf bold}, and the second-best results are \underline{underlined}. Same for the other tables.
    }
    \label{tab:isic17}
    \scriptsize
    \resizebox{\columnwidth}{!}
    {
    \begin{tabular}{l|l|l|l|l}
        \hline
        Method &Acc ($\uparrow$)  &AUC ($\uparrow$) &Sen ($\uparrow$)  &Spe ($\uparrow$)\\\hline
        Galdran et al.~\cite{galdran2017data} &48.00	&76.50	&\textbf{90.60}	&37.70\\
        Vasconcelos et al.~\cite{vasconcelos2017increasing} &83.00	&79.10	&17.10	&\underline{99.00}\\
        D{\'\i}az~\cite{diaz2017incorporating} &82.30	&\textbf{85.60}	&10.30	&\textbf{99.80}\\
        Zhang et al.~\cite{zhang2019medical}  &83.00	&83.00	&---	&---\\
        Suraj et al.~\cite{mishra2022data} &\underline{86.00}	&83.10	&\underline{59.00}	&92.60\\
        ResNet152~\cite{he2016deep}&84.53	&81.28	&49.23	&93.08\\
        SENet154~\cite{hu2018squeeze}& 84.45	&79.41	&42.74	&94.53\\\hline
          ResNet152 + SAMAug-C& 85.07	&81.83	&48.89	&94.20\\
          ResNet152 + SAMAug-C + Ensemble  & 85.67	&\underline{83.93}	&48.72	&94.62 \\\hline
        SENet154 + SAMAug-C& 85.83	&80.82	&44.79	&95.78\\
        SENet154 + SAMAug-C + Ensemble& \textbf{86.67}	&83.27	&47.01	&96.27\\\hline
    \end{tabular}
}
\vspace{-6mm}
\end{table}

\begin{table}[t]
    \centering
    \caption{Results on the vitiligo (public) dataset. 
    }
    \label{tab:vitiligo}
    \scriptsize
    \resizebox{\columnwidth}{!}
    {
    \begin{tabular}{l|l|l|l|l}
        \hline
        Method &Acc ($\uparrow$)  &AUC ($\uparrow$) &Sen($\uparrow$)  &Spe ($\uparrow$)\\\hline
        VGG13~\cite{simonyan2014very}&---	&\underline{0.995}	&0.972	&0.963\\
        ResNet18~\cite{he2016deep}& ---	&0.958	& 0.952 	& 0.957\\
        DenseNet121~\cite{huang2017densely}& ---	&0.982	& 0.962	&0.961\\
        Dermatologists~\cite{zhang2021design}& ---	&---	&0.964	&0.803\\
         Suraj et al.~\cite{mishra2022data} &\underline{0.988}	&\textbf{0.998}	&\underline{0.996}	&\textbf{0.975}\\
         ResNet18 + SAMAug~\cite{zhang2023input} &0.967	&0.992	&0.990	&0.934 \\
        DenseNet121 + SAMAug~\cite{zhang2023input}&0.975	&0.993	&0.983	&0.957 \\
        \hline
         ResNet18 + SAMAug-C& 0.968	&0.993	&0.991	&0.955\\
          ResNet18 + SAMAug-C + Ensemble  & 0.982	&\textbf{0.998}	& \underline{0.996}	&\underline{0.966} \\\hline
         DenseNet121 + SAMAug-C& 0.978	&0.993	&0.993	&0.962\\
          DenseNet121 + SAMAug-C + Ensemble  & \textbf{0.990}	 &\textbf{0.998}	 &\textbf{0.997}	 &\textbf{0.975} \\\hline
          
    \end{tabular}
}
\vspace{-6mm}
\end{table}
\begin{table}[t]
    \centering
    \caption{Results on the ExtCRC dataset. 
    }
    \label{tab:extcrc}
    \scriptsize
    \resizebox{\columnwidth}{!}
    {
    \begin{tabular}{l|l|l|l|l}
        \hline
        Method &Acc ($\uparrow$)  &AUC ($\uparrow$) &Sen ($\uparrow$)  &Spe ($\uparrow$)\\\hline
        ResNet50~\cite{he2016deep}&77.00	&89.07	&74.06	&87.93\\
        ResNeXt50~\cite{xie2017aggregated}& 80.00	& 90.93	&77.28	&89.40\\
        SE-ResNet50~\cite{hu2018squeeze}& 81.33	&90.85	& 78.76	& 90.36 \\
        ResNet50 + SAMAug~\cite{zhang2023input}& 76.67	&88.73	&71.63	&87.52\\
        ResNeXt50 + SAMAug~\cite{zhang2023input}& 76.00	&88.64	&71.67	&87.16\\
        SE-ResNet50 + SAMAug~\cite{zhang2023input}& 81.25	&90.71	&77.44	&89.86\\
        \hline
          ResNet50 + SAMAug-C& 79.67	&90.85	&76.92	&89.41\\
          ResNet50 + SAMAug-C + Ensemble  & 80.67	&91.68	&78.06	&89.70 \\\hline
        ResNeXt50 + SAMAug-C&81.67	&90.80	&79.19	&90.42\\
        ResNeXt50 + SAMAug-C + Ensemble&82.46	&91.67	&80.62	&90.87\\\hline
        SE-ResNet50 + SAMAug-C& \underline{82.67}	& \underline{92.02}	& \underline{81.06}	& \underline{91.19} \\
        SE-ResNet50 + SAMAug-C + Ensemble& \textbf{83.52}	&\textbf{92.69}	&\textbf{82.18}	&\textbf{91.82}\\\hline
    \end{tabular}
}
\vspace{-7.5mm}
\end{table}
\begin{table}[t]
    \centering
    \caption{Results of various ensemble methods on the ISIC 2017 dataset. }
    \label{tab:aba}
    \scriptsize
    \resizebox{\columnwidth}{!}
    {
    \begin{tabular}{l|l|l|l|l}
        \hline
            &Acc ($\uparrow$)  &AUC ($\uparrow$) &Sen ($\uparrow$)  &Spe ($\uparrow$)\\\hline
        \multicolumn{1}{c}{ResNet152 + SAMAug-C}\\\hline
        Voting  & 84.83	&82.29	&49.57	&93.37 \\
        Entropy  & 82.67	&81.89	&38.46	&93.37 \\
        Direct Average  & 85.17	&83.33	&50.43	&93.58 \\
        Weighted Average w/ weights [0.6, 0.4]  & \underline{85.33}	&\underline{84.18}	&50.43	&\underline{93.79} \\
        Weighted Average w/ weights [0.4, 0.6]  & 84.67	 &81.15	 &\underline{55.56}	 &91.72 \\
        Weighted Average w/ weights [0.7, 0.3]  & 84.17	  &\textbf{84.30}	&\textbf{59.83}	&90.06 \\
        Weighted Average w/ weights [0.3, 0.7]  & \textbf{85.67}	&83.93	&48.72	&\textbf{94.62} \\ \hline
        \multicolumn{1}{c}{SENet154 + SAMAug-C}\\\hline
        Voting  & 86.00	&80.78	&\textbf{48.72}	&95.03 \\
        Entropy  & 84.84	&82.18	&45.30	&94.41 \\
        Direct Average  & \underline{86.50}	&83.30	&47.01	&\underline{96.07} \\
        Weighted Average w/ weights [0.6, 0.4]  & 86.33	&81.05	&\underline{47.86}	&95.65 \\
        Weighted Average w/ weights [0.4, 0.6]  & \underline{86.50}	&\underline{83.35}	&\underline{47.86}	&95.86 \\
        Weighted Average w/ weights [0.7, 0.3]  & 86.17	&81.12	&\textbf{48.72}	&95.24 \\
        Weighted Average w/ weights [0.3, 0.7]  & \textbf{86.67}	&\textbf{83.37}	&47.01	&\textbf{96.27} \\\hline
    \end{tabular}
}
\vspace{-8mm}
\end{table}
\noindent
\textbf{Experimental Results.}
\textbf{(1) ISIC 2017 Results.}
To evaluate our method, we use two prominent models: ResNet152 and SENet154,
which were pre-trained on ImageNet and have proven their efficacy on medical image datasets, with teams using them to obtain top scores in the ISIC skin lesion classification challenge.
From Table~\ref{tab:isic17}, we observe that:
(I) The models trained with SAMAug-C augmented images demonstrate superior performance over their counterparts trained solely on raw images. 
Specifically, ResNet152's accuracy is improved by 0.54\%, while SENet154's is improved by 1.38\%. 
This validates the effectiveness of our SAMAug-C augmentation method in boosting medical image classification.
(II) When the models are concurrently trained with both raw and SAMAug-C augmented images, their performances are further bolstered. 
ResNet152's accuracy is lifted by an additional 0.6\%, while SENet154's grows by 0.84\%. 
This finding shows our proposed framework's capability to leverage raw and SAMAug-C augmented images to elevate classification outcomes.
(III) Our method outperforms the SOTA methods (Suraj et al.~\cite{mishra2022data} and Zhang et al.~\cite{zhang2019medical}) in accuracy.
\textbf{(2) Vitiligo (Public) Results.}
%
From Table~\ref{tab:vitiligo}, we observe that:
(I) On employing SAMAug-C augmented images for training, both our baseline models, ResNet18 and DenseNet121, exhibit superior results. 
Specifically, there is an improvement in AUC by 3.50\% with ResNet18 and 1.10\% with DenseNet121.
This further attests to the capability of our SAMAug-C augmentation method.
(II) When the models are simultaneously trained on raw and SAMAug-C augmented images, the AUC results for ResNet18 and DenseNet121 attain an uplift by an additional 0.50\%. 
This indicates that our dual training approach can enhance performance over using just augmented or raw images alone.
(III) Our method slightly outperforms the SOTA method, Suraj et al.~\cite{mishra2022data}, in both accuracy and sensitivity.
These results demonstrate our method's effectiveness.
\textbf{(3) ExtCRC Results.}
In the experiments, we use non-pre-trained versions of three representative (ResNet50, ResNeXt50, and SE-ResNet50) models for gauging the robustness and stability of our method. 
From Table~\ref{tab:extcrc}, we observe that:
(I) Augmenting raw images using our SAMAug-C method leads to a noticeable performance improvement for all three non-pre-trained models.
Interestingly, when these models are trained using augmented raw images generated by the SAMAug~\cite{zhang2023input}, a decline in their performance is evident. 
This observation highlights the superiority and appropriateness of our SAMAug-C augmentation method for medical image classification.
(II) We observe further improvements in model accuracy by simultaneously training on both raw images and their SAMAug-C augmented images. 
Concretely, ResNet50, ResNeXt50, and SE-ResNet50 are improved by 1.0\%, 0.79\%, and 0.85\% in accuracy, respectively. 
These results are a testimony to the potential of our method. 

\noindent
\textbf{Model Ensemble Exploration.}
We conduct experiments to explore different ensemble methods (i.e., voting, entropy, direct average, and weighted average with various weights) using the ISIC 2017 dataset.
As shown in Table~\ref{tab:aba}, for both ResNet152 and SENet154, the weighted average method yields the best accuracy results. The direct average method gives the second-best results, followed by voting. The entropy method attains the least accurate results.
The combination of weights [0.3, 0.7] produces the highest accuracy among the different weight choices for the weighted average method. This suggests that predictions by the model trained with SAMAug-C augmented images are more reliable.

\vspace{-4mm}
\section{Conclusions} \label{concl}
\vspace{-2mm}
In this paper, we presented a new augmentation method (SAMAug-C) that leverages the SAM for augmenting raw image input to improve medical image classification.
To further enhance classification performance, we designed a novel framework that effectively uses both raw and SAMAug-C augmented images.
Experiments on three public datasets demonstrated the efficacy of our new approach.

\vspace{-2mm}
\section{Compliance with ethical standards}
\label{sec:ethics}
\vspace{-2mm}
This research study was conducted retrospectively using human subject data made available in open access by three publicly available datasets~\cite{codella2018skin,zhang2021design,shaban2020context}. Ethical approval was not required as confirmed by the licenses attached with the open access datasets.

\section{Acknowledgements}
\label{sec:acknowledgements}
This research was supported in part by NSF grants IIS-1955395, IIS-2101696, and OAC-2104158.

\small
\bibliographystyle{IEEEbib_abbrev}
\bibliography{refs}
\end{document}